\documentclass{article}
\usepackage{fix-cm}

\usepackage{conference,times}
\usepackage{amsmath,amssymb}
\DeclareFontFamily{OML}{playcraftletters}{\skewchar\font127}
\DeclareFontShape{OML}{playcraftletters}{m}{it}{<->cmmi10}{}
\DeclareFontShape{OML}{playcraftletters}{b}{it}{<->cmmib10}{}
\DeclareFontFamily{OT1}{playcraftoperators}{}
\DeclareFontShape{OT1}{playcraftoperators}{m}{n}{<->cmr10}{}
\DeclareFontShape{OT1}{playcraftoperators}{b}{n}{<->cmbx10}{}
\DeclareFontFamily{OMS}{playcraftsymbols}{\skewchar\font48}
\DeclareFontShape{OMS}{playcraftsymbols}{m}{n}{<->cmsy10}{}
\DeclareFontShape{OMS}{playcraftsymbols}{b}{n}{<->cmbsy10}{}

\DeclareMathVersion{playcraft}
\SetSymbolFont{letters}{playcraft}{OML}{playcraftletters}{m}{it}
\SetSymbolFont{operators}{playcraft}{OT1}{playcraftoperators}{m}{n}
\SetSymbolFont{symbols}{playcraft}{OMS}{playcraftsymbols}{m}{n}

\DeclareMathVersion{playcraftbold}
\SetSymbolFont{letters}{playcraftbold}{OML}{playcraftletters}{b}{it}
\SetSymbolFont{operators}{playcraftbold}{OT1}{playcraftoperators}{b}{n}
\SetSymbolFont{symbols}{playcraftbold}{OMS}{playcraftsymbols}{b}{n}

\newcommand{\playcraftBoldSymbol}[1]{{\mathversion{playcraftbold}$#1$}}

\newcommand{\playcraftLoop}{%
\begingroup%
\mathversion{playcraft}%
\pgfmathsetlengthmacro{\playcraftUnit}{\linewidth/1572}%
\pgfmathsetlengthmacro{\playcraftArrowLength}{17.5*\playcraftUnit}%
\pgfmathsetlengthmacro{\playcraftArrowWidth}{16.45*\playcraftUnit}%
\pgfmathsetlengthmacro{\playcraftSmallArrowLength}{16.4*\playcraftUnit}%
\pgfmathsetlengthmacro{\playcraftSmallArrowWidth}{15.4*\playcraftUnit}%
\pgfmathsetlengthmacro{\playcraftAxisLength}{8.65*\playcraftUnit}%
\pgfmathsetlengthmacro{\playcraftAxisWidth}{8.13*\playcraftUnit}%
\newcommand{\playcraftFont}[1]{\fontsize{\dimexpr\playcraftUnit*##1\relax}{\dimexpr\playcraftUnit*##1*28/25\relax}\selectfont}%
\definecolor{playcraftText}{HTML}{263642}%
\definecolor{playcraftGrayFill}{HTML}{F2F4F6}%
\definecolor{playcraftGrayBorder}{HTML}{83909B}%
\definecolor{playcraftBlueFill}{HTML}{F1F7FD}%
\definecolor{playcraftBlueHeader}{HTML}{DCEAF8}%
\definecolor{playcraftBlueBorder}{HTML}{5080AC}%
\definecolor{playcraftBlueText}{HTML}{174979}%
\definecolor{playcraftPlayerBorder}{HTML}{A4BED8}%
\definecolor{playcraftOrangeFill}{HTML}{FFF7EF}%
\definecolor{playcraftOrangeHeader}{HTML}{F8E5D2}%
\definecolor{playcraftOrangeBorder}{HTML}{BD783F}%
\definecolor{playcraftOrangeText}{HTML}{834110}%
\definecolor{playcraftCrafterBorder}{HTML}{DCB697}%
\definecolor{playcraftArrow}{HTML}{334451}%
\definecolor{playcraftBlueArrow}{HTML}{426C94}%
\definecolor{playcraftCurrentCurve}{HTML}{526170}%
\definecolor{playcraftExpectedCurve}{HTML}{A34E16}%
\definecolor{playcraftAxis}{HTML}{8896A2}%
\definecolor{playcraftNoteBorder}{HTML}{586977}%
\definecolor{playcraftNoteFold}{HTML}{EEF1F4}%
\definecolor{playcraftLabelFill}{HTML}{F9F7ED}%
\definecolor{playcraftLabelBorder}{HTML}{36393D}%
\definecolor{playcraftWorkflowFill}{HTML}{F9E9D9}%
\hyphenpenalty=10000\relax%
\exhyphenpenalty=10000\relax%
\begin{tikzpicture}[
  x=\playcraftUnit,
  y=-\playcraftUnit,
  every node/.style={font=\playcraftFont{22}, text=playcraftText, inner sep=0pt, outer sep=0pt},
  playcraft border/.style={line width=\dimexpr\playcraftUnit*3/2\relax},
  playcraft thin/.style={line width=\playcraftUnit},
  playcraft card/.style={playcraft border, draw=playcraftGrayBorder, fill=playcraftGrayFill},
  playcraft arrow/.style={
    -{Triangle[length=\playcraftArrowLength,width=\playcraftArrowWidth]},
    draw=playcraftArrow, line width=\dimexpr\playcraftUnit*5\relax
  },
  playcraft secondary arrow/.style={
    -{Triangle[length=\playcraftSmallArrowLength,width=\playcraftSmallArrowWidth]},
    draw=playcraftArrow, line width=\dimexpr\playcraftUnit*4\relax
  },
  playcraft blue arrow/.style={playcraft secondary arrow, draw=playcraftBlueArrow},
  playcraft axis/.style={
    -{Triangle[length=\playcraftAxisLength,width=\playcraftAxisWidth]},
    draw=playcraftAxis, line width=\dimexpr\playcraftUnit*3/2\relax
  }
]
  \path[use as bounding box] (10,10) rectangle (1582,600);

  \begin{scope}[shift={(20,110)}]
    \node[playcraft card, minimum width=\dimexpr\playcraftUnit*326\relax,
      minimum height=\dimexpr\playcraftUnit*384\relax] (current-game) at (163,192) {};
    \node[font=\playcraftFont{24}\bfseries] at (163,29) {Current {\playcraftFont{26}\playcraftBoldSymbol{G_k}}};
    \node[anchor=north west] at (16,55) {
      \includegraphics[width=\dimexpr\playcraftUnit*294\relax,height=\dimexpr\playcraftUnit*441/2\relax]{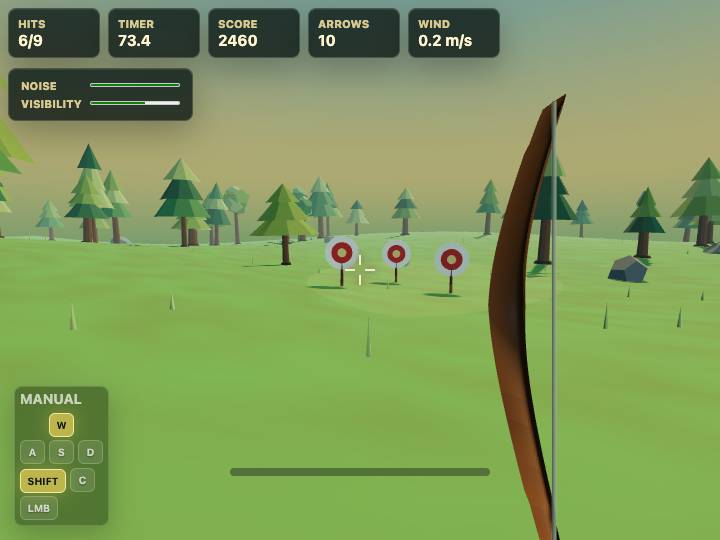}
    };
    \node[align=center, font=\playcraftFont{24}\itshape] at (163,318) {
      Approaching targets\\and aiming each shot
    };
  \end{scope}

  \begin{scope}[shift={(471,63)}]
    \path[playcraft border, draw=playcraftBlueBorder, fill=playcraftBlueFill]
      (0,0) rectangle (316,466);
    \path[playcraft border, draw=playcraftBlueBorder, fill=playcraftBlueHeader]
      (0,0) rectangle (316,49);
    \node[anchor=west, font=\playcraftFont{26}\bfseries, text=playcraftBlueText]
      at (17,24.5) {Playing stage};
    \node[anchor=west, font=\playcraftFont{26}\bfseries] at (18,76) {Player agent};

    \node[playcraft thin, draw=playcraftPlayerBorder, fill=white,
      minimum width=\dimexpr\playcraftUnit*274\relax,
      minimum height=\dimexpr\playcraftUnit*165\relax] (player) at (158,183.5) {};
    \node[anchor=west, align=left, font=\playcraftFont{22}] at (39,183.5) {
      Pursue the game objective\\
      across $E$ episodes while\\
      carrying memory forward.\\[\dimexpr\playcraftUnit*22\relax]
      Review each episode\\
      and write a report.
    };

    \begin{scope}[shift={(11,316)}]
      \node[playcraft thin, draw=playcraftNoteBorder, fill=white,
        rounded corners=\dimexpr\playcraftUnit*39/2\relax,
        minimum width=\dimexpr\playcraftUnit*294\relax,
        minimum height=\dimexpr\playcraftUnit*130\relax] (play-record) at (147,65) {};
      \node[font=\playcraftFont{22}\bfseries] at (147,20.5) {Play record};
      \foreach \reportX/\reportNumber in {11/1,99/2,203/{$E$}} {
        \begin{scope}[shift={(\reportX,43)}]
          \path[playcraft thin, draw=playcraftNoteBorder, fill=white]
            (0,0) -- (69,0) -- (80,11) -- (80,73) -- (0,73) -- cycle;
          \path[playcraft thin, draw=playcraftNoteBorder, fill=playcraftNoteFold]
            (69,0) -- (69,11) -- (80,11);
          \node[align=center, font=\playcraftFont{20}] at (40,38.5) {Report\\\reportNumber};
        \end{scope}
      }
      \node[font=\playcraftFont{22}, text=playcraftCurrentCurve] at (190.5,79.5) {...};
    \end{scope}
  \end{scope}

  \begin{scope}[shift={(828,63)}]
    \path[playcraft border, draw=playcraftOrangeBorder, fill=playcraftOrangeFill]
      (0,0) rectangle (384,466);
    \path[playcraft border, draw=playcraftOrangeBorder, fill=playcraftOrangeHeader]
      (0,0) rectangle (384,48);
    \node[anchor=west, font=\playcraftFont{26}\bfseries, text=playcraftOrangeText]
      at (17,24) {Crafting stage};
    \node[anchor=west, font=\playcraftFont{26}\bfseries] at (15,76) {Crafter agent};

    \begin{scope}[shift={(13,101)}]
      \node[playcraft thin, draw=playcraftCrafterBorder, fill=white,
        minimum width=\dimexpr\playcraftUnit*358\relax,
        minimum height=\dimexpr\playcraftUnit*344\relax] (crafter) at (179,172) {};
      \node[anchor=west, font=\playcraftFont{23}\bfseries] at (18,26) {Design targets};
      \node[anchor=west, align=left, font=\playcraftFont{22}] at (18,76) {
        \textbullet\ Meet performance anchors\\
        \textbullet\ Reward greater skill
      };

      \begin{scope}[shift={(16,125)}]
        \node[anchor=west, font=\playcraftFont{23}\bfseries] at (2,13.716) {Learning curves};
        \node[anchor=west, font=\playcraftFont{19}, text=playcraftCurrentCurve]
          at (2,42.4704) {Player Performance};
        \draw[playcraft axis] (4.82,129.11) -- (4.82,63.46154);
        \draw[playcraft axis] (4.82,129.11) -- (182.57651,129.11);

        \draw[draw=playcraftCurrentCurve, line width=\dimexpr\playcraftUnit*5/2\relax]
          (6,104.96119)
          .. controls (24,101.18233) and (39.16667,98.51183) .. (51.5,96.94970)
          .. controls (63.83333,95.38757) and (77.83333,93.88240) .. (93.5,92.43417)
          .. controls (109.16667,90.98595) and (121.66667,89.97707) .. (131,89.40754)
          .. controls (140.33333,88.83802) and (151.66667,88.36504) .. (165,87.98862);
        \draw[draw=playcraftExpectedCurve, line width=\dimexpr\playcraftUnit*5/2\relax]
          (5.16726,120.76468)
          .. controls (18.38909,117.31998) and (29.83333,114.14941) .. (39.5,111.25296)
          .. controls (49.16667,108.35651) and (60.33333,104.59763) .. (73,99.97633)
          .. controls (85.66667,95.35503) and (98.5,90.29438) .. (111.5,84.79438)
          .. controls (124.5,79.29438) and (134.66667,75.07988) .. (142,72.15089)
          .. controls (149.33333,69.22189) and (157.66667,66.99118) .. (167,65.45875);
        \node[font=\playcraftFont{19}, text=playcraftCurrentCurve] at (93.44128,146.44970) {Play experience};
      \end{scope}

      \node[anchor=west, font=\playcraftFont{20}\bfseries] at (198,212.9886) {Current};
      \node[anchor=west, font=\playcraftFont{20}\bfseries, text=playcraftExpectedCurve]
        at (198,190.4587) {Expected};
      \path[fill=playcraftWorkflowFill, rounded corners=\dimexpr\playcraftUnit*32/5\relax]
        (8,299) rectangle (350,331);
      \node[font=\playcraftFont{20}\bfseries, text=playcraftOrangeText] at (179,315) {
        Diagnose $\rightarrow$ Propose $\rightarrow$ Implement
      };
    \end{scope}
  \end{scope}

  \begin{scope}[shift={(1246,110)}]
    \node[playcraft card, minimum width=\dimexpr\playcraftUnit*326\relax,
      minimum height=\dimexpr\playcraftUnit*384\relax] (revised-game) at (163,192) {};
    \node[font=\playcraftFont{24}\bfseries] at (163,29) {Revised {\playcraftFont{26}\playcraftBoldSymbol{G_{k+1}}}};
    \node[anchor=north west] at (16,55) {
      \includegraphics[width=\dimexpr\playcraftUnit*294\relax,height=\dimexpr\playcraftUnit*441/2\relax]{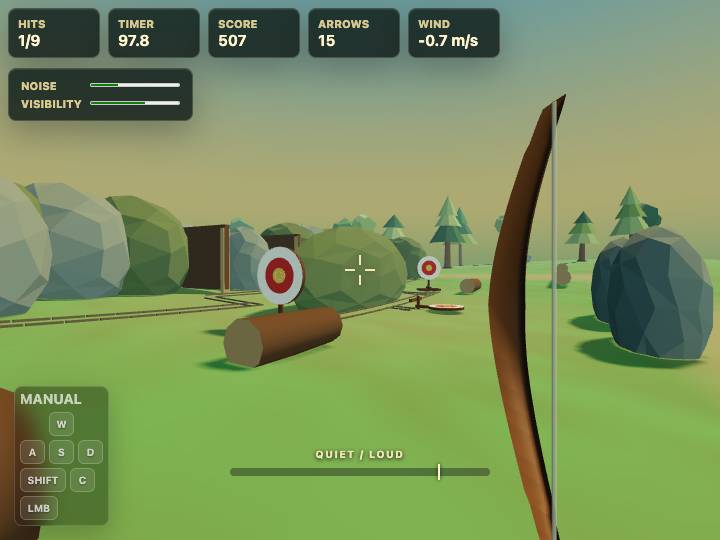}
    };
    \node[align=center, font=\playcraftFont{24}\itshape] at (163,318) {
      Timing quiet shots\\before targets take cover
    };
  \end{scope}

  \node[playcraft card, minimum width=\dimexpr\playcraftUnit*326\relax,
    minimum height=\dimexpr\playcraftUnit*70\relax] (initial-game) at (183,49) {};
  \node[align=center, font=\playcraftFont{24}] at (183,49) {
    {\bfseries Initial game {\playcraftFont{26}\playcraftBoldSymbol{G_0}}}\\
    {\playcraftFont{22}Prompt-generated or existing}
  };

  \coordinate (image-flow) at (0,224);
  \coordinate (action-flow) at (0,257);
  \coordinate (record-entry) at (0,299);
  \coordinate (revision-flow) at (0,302);
  \draw[playcraft arrow] (initial-game.south) -- (current-game.north);
  \draw[playcraft blue arrow] (current-game.east |- image-flow) -- (player.west |- image-flow);
  \draw[playcraft blue arrow] (player.west |- action-flow) -- (current-game.east |- action-flow);
  \draw[playcraft secondary arrow] (player.south) -- (play-record.north);
  \draw[playcraft arrow] (play-record.east) -- (807,444) -- (807,299) -- (crafter.west |- record-entry);
  \draw[playcraft arrow] (crafter.east |- revision-flow) -- (revised-game.west |- revision-flow);
  \draw[playcraft arrow] (revised-game.south) -- (1409,559) -- (183,559) -- (current-game.south);

  \foreach \labelX/\labelY/\labelName in {408.6274/189/Images,411/295/Actions} {
    \node[playcraft thin, draw=playcraftLabelBorder, fill=playcraftLabelFill,
      rounded corners=\dimexpr\playcraftUnit*15/2\relax,
      minimum width=\dimexpr\playcraftUnit*98\relax,
      minimum height=\dimexpr\playcraftUnit*48\relax,
      font=\playcraftFont{24}] at (\labelX,\labelY) {\labelName};
  }
  \node[fill=white, font=\playcraftFont{25}\bfseries] at (796,576) {Next Iteration};
\end{tikzpicture}%
\endgroup%
}

\usepackage{microtype}
\usepackage{graphicx}
\usepackage{pgfplots}
\usepackage{pgfplotstable}
\pgfplotsset{compat=1.18}
\usepgfplotslibrary{groupplots}
\usetikzlibrary{arrows.meta}
\usepackage{booktabs}
\usepackage[section]{placeins}
\usepackage{xcolor}
\usepackage{xurl}
\usepackage[unicode,breaklinks]{hyperref}

\hypersetup{
  colorlinks=true,
  linkcolor=blue!50!black,
  citecolor=blue!50!black,
  urlcolor=blue!50!black,
  pdftitle={PlayCraft: Designing Engaging Games from Agent Play Experience}
}
\title{PlayCraft: Designing Engaging Games\\from Agent Play Experience}
\iclrfinalcopy
\author{%
  Weihao Cheng\\[0.2em]
  {\normalsize weihao.cheng@proton.me}
}
\date{}

\makeatletter
\renewcommand*{\fps@figure}{!tp}
\renewcommand*{\fps@table}{!tp}
\makeatother

\begin{document}
\raggedbottom
\maketitle
\lhead{}

\begin{abstract}
Recent AI agents can generate playable games, but these games are not necessarily engaging.
Mastery is a key source of engagement, yet designing for it requires extensive human play and tuning.
We propose PlayCraft, an automated framework that uses a vision-based agent to play games and draws on its detailed play experience to revise them for mastery.
Starting from an initial game, PlayCraft alternates between playing and crafting.
In the playing stage, a player agent watches the game screen, uses the controls, and pursues the game objective over repeated episodes while retaining memory.
After each episode, it reports its progress, outcomes, decisions, and reflections.
In the crafting stage, a crafter agent uses these reports to revise the game. 
It estimates player abilities, predicts performance at different proficiency levels, and compares those predictions with a target learning progression.
It diagnoses the game and addresses the gaps by giving players meaningful opportunities to improve through better decisions and execution.
We evaluate games in fresh runs with player agents across three intelligence settings, measuring learning gain and learning slope.
Compared with the initial games, PlayCraft yields larger learning gains and steeper learning slopes in most games. 
It also outperforms the baselines on the macro-average of each metric at every intelligence setting.
\end{abstract}

\begin{quote}
  \small
  \noindent Website: \url{https://xt4d.github.io/playcraft-web} \\[0.15em]
  Code: \url{https://github.com/xt4d/PlayCraft}
\end{quote}

\section{Introduction}\label{sec:introduction}

Recent AI agents can generate playable games from human instructions \citep{huang2026}, but making these games engaging remains difficult.
A central source of engagement is mastery, which is the process of developing strategies and skills to overcome the challenges of a game \citep{koster2004}.
A game supports mastery when players initially achieve limited success but reach substantially better results as they become more proficient.
This progression motivates repeated play, but designing for it requires extensive human play and tuning.
We therefore study how to automatically generate games that support mastery.

Prior work generates games that reward stronger play by using play results from computational players to guide designs \citep{togelius2008,nasir2026}.
However, the strategies these players find through search or game-specific training may differ from those people develop, making their behavior less representative of human play.
Other systems use feedback from repeated playtests to revise generated games \citep{huang2026,wu2026rsigame,chen2026recursive}.
These systems prioritize overall game quality, with limited explicit focus on player learning through practice.

Recent agents based on foundation models, such as GPT-6 Astra~\citep{openai2026gpt6astra}, can play 2D and 3D games directly from rendered frames \citep{yu2026lolplay,aggarwal2026gymanythingrt}.
They perceive only what the game shows, act through the same controls as human players, and can learn from earlier experience.
They can also report their understanding of the game and the difficulties they encounter.
These capabilities make them suitable simulated players whose experience can shape challenges that reward skillful play.

We propose PlayCraft, an automated framework that starts from an initial playable game and alternates between playing and crafting stages.
In the playing stage, a foundation-model player agent observes rendered frames and interacts with the game through the provided controls. 
It plays a series of episodes, gaining experience with the mechanics and challenges it encounters.
The game resets between attempts, while the player agent retains its memory and the lessons it has learned.
After each episode, it writes a report covering its progress, outcomes, decisions, and reflections.  
In the crafting stage, a crafter agent uses these reports to diagnose and revise the game to promote engagement through mastery. 
It first estimates the baseline abilities that the player agent has before playing the game.
Using these abilities and the observed play, it estimates player performance in the current game at beginner, intermediate, and expert levels.
It then identifies gaps between these estimates and a target progression, defined by performance anchors relative to the expected expert outcome.
Lastly, it addresses these gaps by revising the game to provide meaningful opportunities for better decisions and execution, while preserving the objectives and core identity of the game.

We conduct experiments covering ten 2D and 3D games with varied execution and planning demands.
The games are evaluated at three agent intelligence settings using the visual play protocol from the playing stage.
We use gains in agent performance across episodes as a proxy for mastery, assuming that the learning patterns of the player agent resemble those of human players.
We compute learning gain and learning slope \citep{li2026} for the initial games, games produced by PlayCraft, and games revised through code analysis alone or playtesting with policy-based controllers.
Compared with the initial games, PlayCraft increases macro-average learning gain by 2.7, 14.0, and 21.9 percentage points at the Low, Medium, and High intelligence settings, respectively.
Its macro-average learning slope increases by 2.09 and 2.64 at Medium and High, respectively, and is nearly tied at Low.
PlayCraft also achieves higher macro-averages for both metrics than either baseline at every intelligence setting.

\section{Related Work}\label{sec:related-work}

\textbf{Mastery and player experience.} Game design theory has long tied engagement to learning.
\citet{koster2004} argues that games are fun because players learn their patterns and develop mastery, and self-determination theory links enjoyment and continued play to perceived competence \citep{ryan2006}.
\citet{linehan2014} analyze how successful puzzle games pace their challenges as players learn.
Experience-driven procedural content generation \citep{yannakakis2011} uses computational models of player experience to guide the search for
game content that elicits a desired experience.
PlayCraft takes mastery as its design goal and uses a surrogate based on gains in player performance across repeated attempts.

\textbf{Automatic game design with computational players.} Automatic playtesting uses computational players to simulate distinct play styles \citep{holmgard2018}, estimate difficulty for human players \citep{gudmundsson2018}, and predict player churn \citep{roohi2021}.
Agent performance can also guide the selection of candidate designs or the tuning of game parameters.
\citet{togelius2008} evolve the rules of small grid games and score each game by how well neural controllers learn to play it, treating learnability as a surrogate for fun.
Later work uses self-play to evaluate board-game quality \citep{browne2010} and exact game-theoretic values for restricted play to assess competitive balance \citep{jaffe2012}. 
Other systems use the win rate of a stronger agent against a weaker one as a proxy for game depth \citep{liu2017,todd2024}.
Mortar \citep{nasir2026} applies this idea to game code, evolving game mechanics through LLM-driven quality-diversity search and rewarding games in which fixed agents (MCTS with different budgets, random, and no-op) rank in their expected order of skill.
Fly, Fail, Fix \citep{zook2025} uses a pretrained DQN to play \emph{Flappy Bird}.
It then uses GPT-4.1~\citep{openai2025gpt41} to adjust the pipe parameters (gap size, spacing, and speed) based on the resulting play metrics and frame strips to bring the score closer to a target.
Computational players that rely on search or game-specific training may behave differently from humans, require substantial computation to learn unfamiliar games, and offer little insight into their decisions or difficulties. 
To better approximate human play, PlayCraft uses a foundation-model agent that infers mechanics from visual cues, draws on prior knowledge to begin playing, and reflects on earlier attempts to improve. 
The agent can also explain its decisions and link difficulties to specific situations in the game.

\textbf{LLM-based game generation and revision.} LLM agents now generate playable games by decomposing prompts into hierarchical plans, coordinating specialized development and review agents, and reusing skills and knowledge from earlier projects \citep{earle2024,yin2026,jiang2026,hu2026}. Benchmarks evaluate the resulting games through multimodal judging and interaction-based tests \citep{zhang2025vgame,luo2026,chen2026gamexpert}.
Several recent systems use playtest feedback to refine the games they generate.
Play2Code \citep{huang2026} alternates HTML game generation with playtesting by a rubric-blind GUI agent that returns gameplay traces and actionable feedback.
RSIGame \citep{wu2026rsigame} uses targeted playtests to diagnose issues and verify edits, while a separate monitor compares builds and retains the one it judges best.
Recursive Game Creator \citep{chen2026recursive} organizes recursive refinement around Designer, Builder, Player, and Reviewer roles.
Its Player writes reusable policies that act through programmatic interfaces, and its Reviewer uses the resulting trajectories to infer player preferences and recommend changes.
These systems aim to enhance overall game quality.
PlayCraft focuses on mastery, revising games so that players can improve their performance through practice.

\textbf{Visual game-playing agents.} Agents can play commercial video games through the same interface as human players, choosing keyboard and mouse actions based on screen observations \citep{tan2024,sima2024}.
Early benchmarks found that agents often struggled to use those observations to make sustained progress toward game objectives \citep{paglieri2025,zhang2025videogame}.
More recent benchmarks demonstrate the visual game-playing capabilities of GPT-6 Astra~\citep{openai2026gpt6astra}.
On LOLPlay, it completed 9 of 36 \emph{League of Legends} tasks using screenshots \citep{yu2026lolplay}.
In the online track of GameHorizon-Bench, it completed 9 of 20 multi-step tasks in \emph{Minecraft} \citep{wang2026gamehorizon}.
On Gym-Anything-RT, it passed 9\% of the minigames in real time and 75\% with the games paused between decisions \citep{aggarwal2026gymanythingrt}.
PlayCraft draws on these advances in visual gameplay by using a foundation-model agent that observes rendered frames and acts through the same controls available to human players.

\textbf{Agent learning from experience.} Agents can draw on earlier play when facing a new attempt or task. Reflexion carries lessons from completed trials into later attempts \citep{shinn2023}, and Voyager stores acquired skills for reuse in new tasks \citep{wang2023}.
Learn2Play Bench measures agent improvement over repeated attempts at text games with unfamiliar rules, using metrics such as the learning gain between early and late episodes \citep{li2026}.
PlayCraft uses experience from the player agent to guide game design and assesses support for mastery through performance gains across episodes.

\section{Methodology}\label{sec:methodology}

\subsection{Foundation Models as Players}\label{sec:foundation-models-as-players}

Recent AI agents based on foundation models such as GPT-6 can play both 2D and 3D games like human players.
Given a brief game instruction, they infer the mechanics from screenshots, interact through the provided controls, observe the outcomes, and improve their performance with continued play.
Three properties make them suitable as automatic players:

\begin{itemize}
\item They can draw on prior knowledge of game genres, controls, and conventions to begin playing an arbitrary game.
\item They can play from rendered frames and use the same controls as human players, so their performance can reflect the visual cues and control demands that shape human play.
\item They can learn across attempts by testing hypotheses about rules and strategies, then carrying the lessons into later episodes. They can also report on their play experience, describing the situations they encounter and sharing their perspective on the game.

\end{itemize}
Together, these capabilities let a foundation-model agent learn through repeated play and provide valuable information for game design.

\subsection{PlayCraft}\label{sec:playcraft}

\subsubsection{Overview}\label{sec:overview}

\begin{figure}
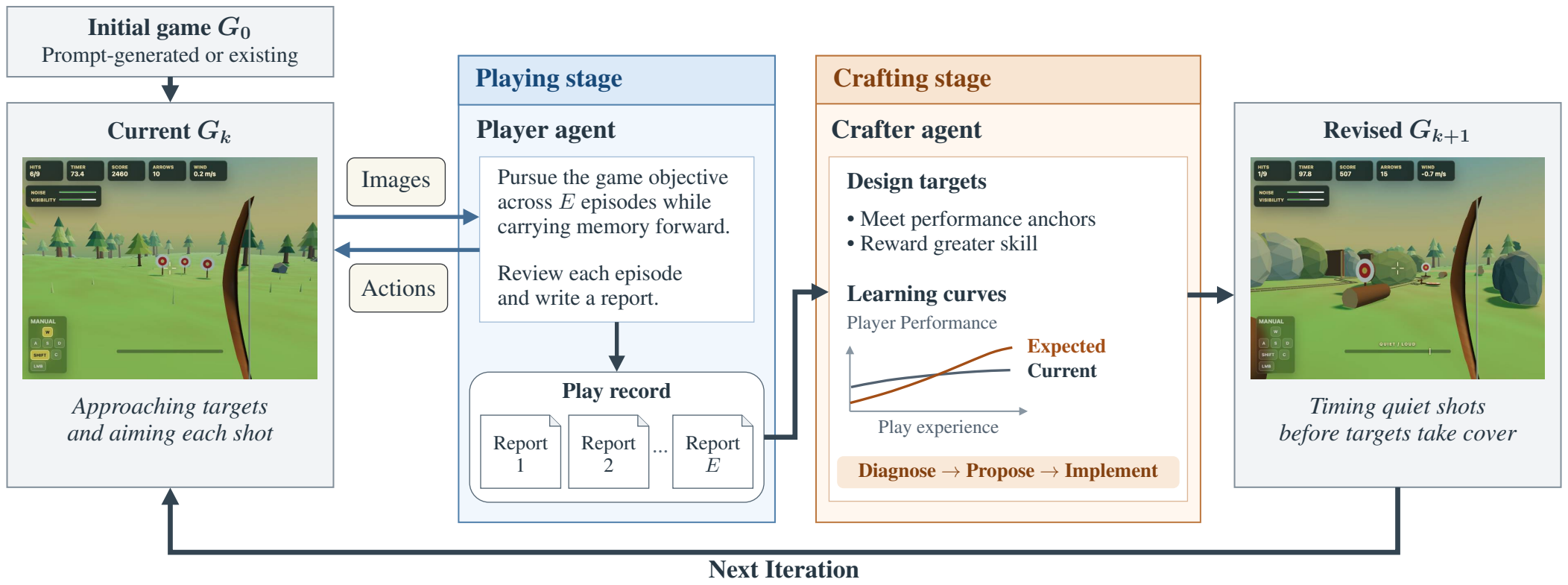

  \centering
  \playcraftLoop
  \caption{Overview of PlayCraft.
  PlayCraft starts from an initial game $G_0$. Each iteration $k$ has a playing stage and a crafting stage.
  In the playing stage, the player agent receives a brief game instruction and plays the current game $G_k$ for $E$ episodes.
  Within an episode, it observes recent frame images and returns control actions.
  After each episode, it reviews its play, writes a report, and carries memory into the next episode.
  The $E$ reports form the final play record.
  In the crafting stage, the crafter agent reads the play record.
  It revises $G_k$ to give players opportunities to improve through better decisions and execution, aiming to meet the predefined performance anchors.
  It produces $G_{k+1}$ for the next iteration.
  The illustrative learning curves show the intended increase in room for mastery.}
  \label{fig:playcraft-loop}
\end{figure}

PlayCraft is an automated framework that designs games for mastery. It starts from an initial playable game $G_0$ and produces revised games $G_1, \dots, G_K$.
Each play--craft iteration $k=0,\ldots,K-1$ has two stages.
In the playing stage, the player agent receives a brief game instruction and pursues the game objective over repeated episodes of $G_k$.
It retains its memory across episodes and writes a report after each one. The reports form the play record passed to the crafting stage.
In the crafting stage, the crafter agent uses the play records to diagnose the game.
It estimates baseline player abilities and compares expected performance at different proficiency levels with predefined anchors.
It then proposes and implements changes that address the gaps and give players room to improve through better decisions and execution, producing $G_{k+1}$ for the next iteration.
Figure~\ref{fig:playcraft-loop} illustrates the PlayCraft loop.

\subsubsection{Playing Stage}\label{sec:playing-stage}

\textbf{Setup.} The player agent begins with a fresh context containing a brief game instruction that states the objective, basic rules, and available controls.
The player agent observes only rendered frames and cannot inspect the game code or access the Internet.
It generates every control action through model inference rather than writing a program to control the game.

\textbf{Interaction.} A run consists of $E$ episodes. Each episode ends when the game reaches a defined endpoint or the active play budget $T$ is exhausted.
Game frames are captured at intervals of $O$, and the player agent makes a decision at intervals of $W$.
At each decision, it receives the frames captured during the preceding decision interval and returns an action plan for the next interval of play.
The plan is a sequence of $A$ control actions, each lasting for a duration $W/A$. A control action is a combination of keys, mouse buttons, and mouse movements.
While the plan executes, the game advances and renders frames, which are included in the next decision request so that the player observes the consequences of its previous plan.
Game timers and animations pause while the model generates the next action plan, so decision latency does not consume play time.
With each plan, the player also reports the visible result, such as the current score, and provides brief notes on its observations and choices.

\textbf{Learning across episodes.} The player agent maintains a single conversation throughout a run.
The game resets between episodes, but the player retains its memory.
After each episode, the player reviews its play and writes a report covering (1) the starting conditions; (2) its progress, including milestones and failures; (3) the final result; (4) its key decisions and the alternatives it considered; (5) its perception of the game, including its understanding of the mechanics, points of confusion, perceived difficulty, and discrepancies between expected and observed outcomes; and (6) the lessons it learned.
The player then updates a consolidated set of natural-language lessons, retaining supported lessons, merging overlapping ones, and revising or removing contradicted ones.
It begins the next episode with its memory of earlier attempts and the updated lessons.
The reports of all episodes form the play record of the run, which is passed to the crafting stage.

\subsubsection{Crafting Stage}\label{sec:crafting-stage}

\textbf{Design targets.} The crafter agent revises the game toward two targets.
The first is to ensure that players at each proficiency level can meet the corresponding predefined performance target.
The most skilled players should be able to complete the objective within the active play budget $T$. In games without an objective, they should be able to experience the core loop and major challenges within that budget. Their expected result defines the 100\% reference.
Beginners and intermediate players should achieve $r_0\%$ and $r_1\%$ of the expert reference, respectively, with $r_0 < r_1$.
These results must come from strategies that would remain advantageous without the time limit.
The second is to give players meaningful opportunities to improve through better decisions and execution. 
The game should make the benefits, costs, and risks of different choices clear and allow expected performance to continue rising as players gain skill.

\textbf{Active play budget.} We assume that players have a finite active play budget $T$, which can be set arbitrarily large.
Introducing $T$ provides a reference range for player performance, allowing the crafter agent to identify and address gaps between estimated performance and the anchors.
Changing $T$ can affect the progression of the resulting game because the crafter uses play observed within that budget as evidence for its design.

\textbf{Diagnosis.} The crafter agent reads the game instruction, the play records of all available runs, and the archived history of earlier iterations.
Because $E$ episodes may not suffice for the player agent to reach expert-level performance, the crafter agent cannot determine the 100\% reference directly from the records.
It therefore first estimates the baseline abilities that the player agent has before playing the game, including screen perception, control execution, and strategy discovery.
A player without prior experience of a game is not necessarily unskilled.
From these abilities and the observed play, it estimates the expected results of beginner, intermediate, and expert players in the current game and compares them with the anchors.
Where the estimates deviate from the anchors, it diagnoses the causes, considering entry difficulty; demands on understanding, perception, and execution; opportunities for strategic choice; reward structure; challenge progression; and performance ceiling.
If an earlier proposal exists, it also checks which expectations in that proposal the new records support, contradict, or leave untested.
It then chooses between building on the earlier direction and exploring a new one.

\textbf{Proposal.} The crafter agent then writes a focused design proposal.
It prioritizes distinctive changes or additions that enable new strategies and meaningful choices while preserving the defining interactions of the game.
It keeps the design coherent and simple, weighing the learning benefit of each rule against the complexity it adds. 
It builds challenges on the baseline abilities of the player rather than assuming that unfamiliar mechanics alone will limit beginner performance.
The revised game must remain visually clear, conveying objects, actions, and consequences through presentation and behavior rather than added indicators.
The proposal describes each change and the findings that motivate it. 
It also states the expected experience and results for beginner, intermediate, and expert players, along with the basis and uncertainty of those estimates.

\textbf{Implementation.} In a separate session, the crafter agent implements the proposal.
For each proposed change, the crafter identifies the intended player-facing outcome and specifies observable acceptance checks. 
It then modifies the game code, implementing foundational changes before those that depend on them.
It verifies the result by running builds and automated checks, exercising the affected gameplay, and inspecting the affected visuals in the actual renderer.
It then corrects any discrepancies.
It records deviations from the proposal with their reasons and updates the game instruction to state only the objective, basic rules, and available controls, leaving the rest for players to discover.

\section{Experiment}\label{sec:experiment}

\subsection{Setup}

\begin{figure}
  \centering
  \footnotesize
  \setlength{\tabcolsep}{0.0075\linewidth}
  \begin{tabular}{@{}ccccc@{}}
    \includegraphics[width=0.187\linewidth]{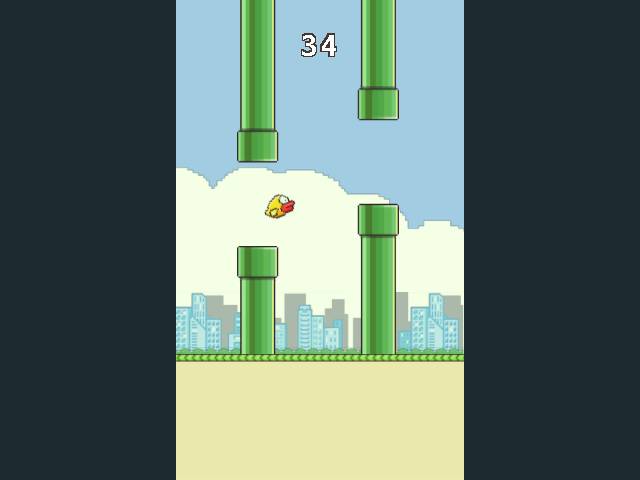} &
    \includegraphics[width=0.187\linewidth]{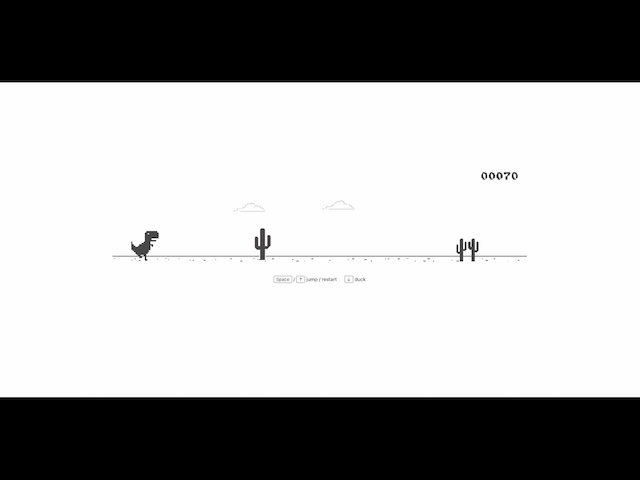} &
    \includegraphics[width=0.187\linewidth]{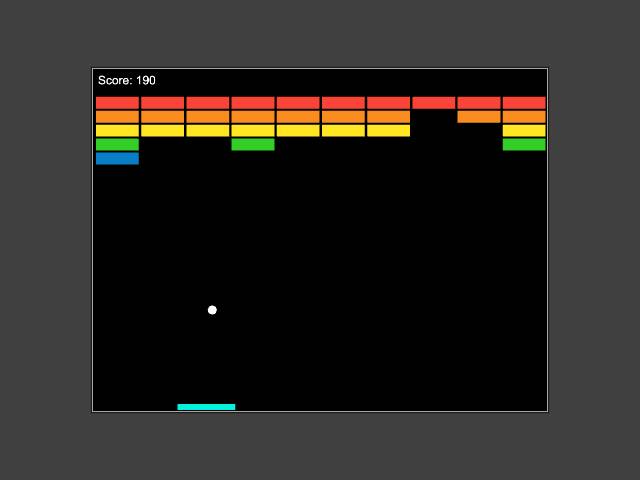} &
    \includegraphics[width=0.187\linewidth]{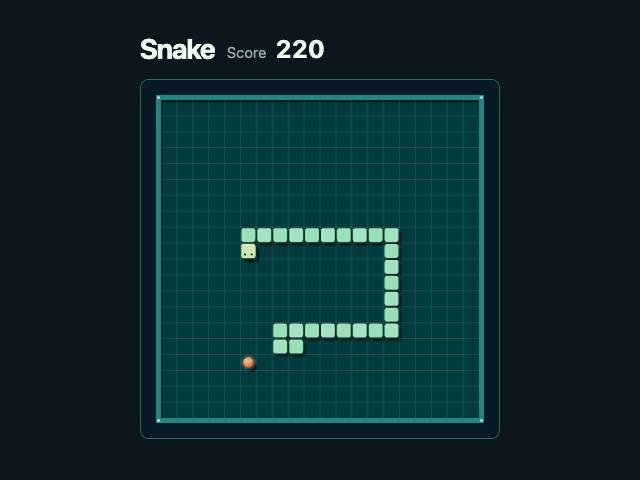} &
    \includegraphics[width=0.187\linewidth]{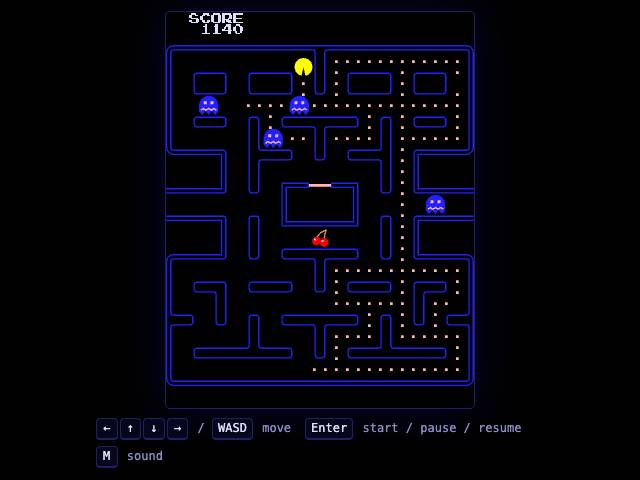} \\
    Flappy Bird & T-Rex Runner & Breakout & Snake & Pac-Man \\[0.7em]
    \includegraphics[width=0.187\linewidth]{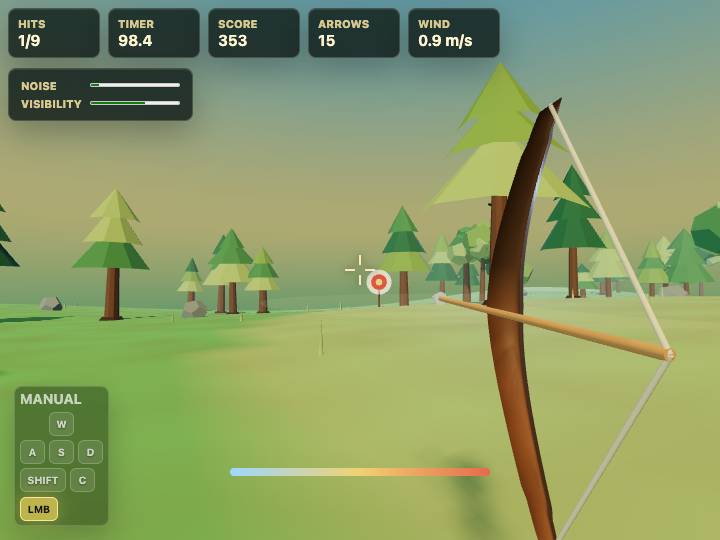} &
    \includegraphics[width=0.187\linewidth]{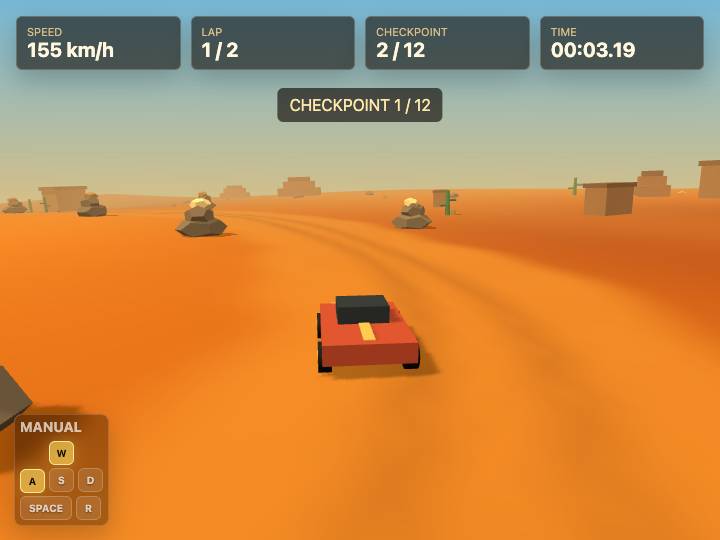} &
    \includegraphics[width=0.187\linewidth]{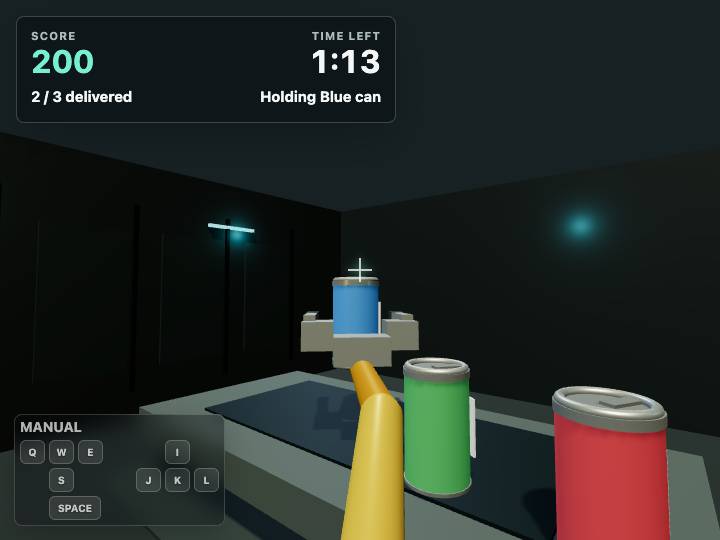} &
    \includegraphics[width=0.187\linewidth]{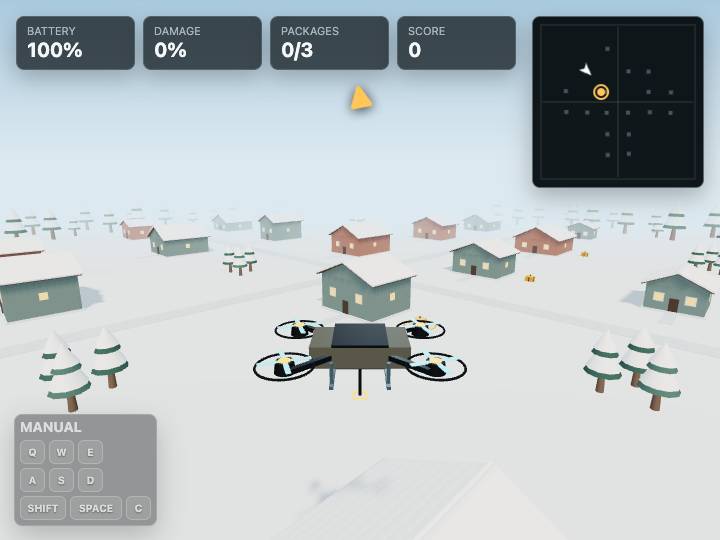} &
    \includegraphics[width=0.187\linewidth]{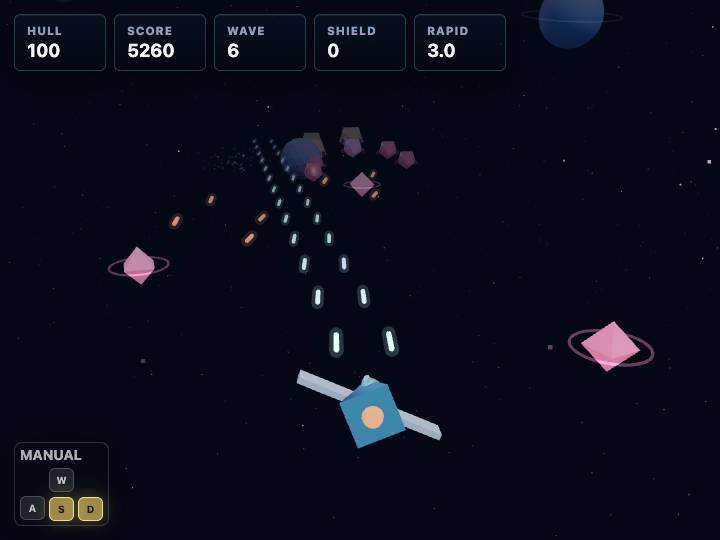} \\
    Archery Hunting & Desert Racing & Robotic Arm & Drone Delivery & Space Shooter
  \end{tabular}
  \caption{Screenshots of the 10 initial games used in the experiments. The top row shows the 2D recreations and the bottom row shows the prompt-generated 3D games.}
  \label{fig:game-screenshots}
\end{figure}

We conduct experiments on 10 browser-based games implemented with Three.js.
The suite includes recreations of 5 classic 2D games: \emph{Flappy Bird}, \emph{T-Rex Runner}, \emph{Breakout}, \emph{Snake}, and \emph{Pac-Man}.
These recreations preserve the core gameplay, with details that may differ from other implementations.
The remaining 5 games are 3D games generated from prompts: \emph{Archery Hunting}, \emph{Desert Racing}, \emph{Robotic Arm}, \emph{Drone Delivery}, and \emph{Space Shooter}.
The suite covers a range of gameplay mechanics, including obstacle avoidance, shooting, racing, navigation, and object manipulation.
Figure~\ref{fig:game-screenshots} shows screenshots of the 10 games.
For each game, we provide the player with brief instructions covering the objective, basic rules, and controls.
We use Codex~0.162.1 to run the foundation-model agents in all experiments.

\subsubsection{PlayCraft setup}

\textbf{Playing.} PlayCraft follows the visual playing protocol in Section~\ref{sec:playing-stage}.
Frames are captured every $O=200$\,ms, and the player makes a decision every $W=1{,}000$\,ms of active play.
Each decision request contains the 5 frames captured during the preceding interval; the first request contains only the initial frame.
The player returns an action plan for the next $1{,}000$\,ms, consisting of $A=50$ control actions lasting for $20$\,ms each.
A control action specifies keys, mouse buttons, and mouse movements.
It may also be empty.
An episode ends when the game reaches an end condition or after $T=60$ seconds of active play.
The game pauses during model inference, so model latency does not consume the active play budget.
Each playing-stage run consists of $E=10$ episodes.
After each episode, the player reports its experience and updates its lessons, which are retained for subsequent episodes.
We use GPT-6.1 Sol~\citep{openai2026} with medium reasoning effort for action planning and xhigh reasoning effort for episode review.

\textbf{Crafting.} The crafter agent follows the diagnosis, proposal, and implementation procedure in Section~\ref{sec:crafting-stage}.
It uses the play records to estimate baseline player abilities and expected performance at beginner, intermediate, and expert levels.
The performance anchors are 10--30\% for beginners and 50--70\% for intermediate players, relative to the expected expert result within the active play budget.
The agent addresses gaps between these estimates and the anchors by designing challenges that reward stronger play, taking baseline player abilities into account.
Design and implementation take place in separate sessions, both using GPT-6.1 Sol, with xhigh reasoning effort for design and max reasoning effort for implementation.
Starting from each initial game, PlayCraft completes 3 play--craft iterations to produce the final game.

\subsubsection{Comparison methods}

We compare PlayCraft with two methods that use different evidence for game revision:
\begin{itemize}
\item \textbf{Code} uses code analysis by a coding agent (Codex), without evidence from a player agent.
\item \textbf{Policy} generates 10 policy-based controllers that act on the internal game state to collect playtesting evidence.
The controllers do not retain experience across attempts.
This method adapts the generated-policy approach of Recursive Game Creator~\citep{chen2026recursive}.
\end{itemize}
Both methods use the same design targets, proposal and implementation procedure, and model settings as PlayCraft, with diagnosis and design guided by the evidence available to each method.
Both methods start from the same initial games as PlayCraft and perform 3 revision iterations.

\subsubsection{Evaluation protocol}

We evaluate the initial games and the final games produced by Code, Policy, and PlayCraft in fresh, independent runs using the visual playing setup described above.
We control player intelligence through 3 combinations of model and reasoning effort, as shown in Table~\ref{tab:player-settings}.
The labels Low, Medium, and High identify these experimental settings.
Action planning and episode review use the same model within each setting.
Each initial or final game is evaluated in 3 independent runs of 10 episodes at each intelligence setting, yielding 9 runs in total.

\begin{table}
\centering
\small
\caption{Player intelligence settings used for evaluation. Episode review includes experience reporting and lesson updates.}
\label{tab:player-settings}
\begin{tabular}{@{}llll@{}}
\toprule
Setting & Model & Action planning & Episode review \\
\midrule
Low & GPT-6.1 Terra & medium & xhigh \\
Medium & GPT-6.1 Sol & medium & xhigh \\
High & GPT-6.1 Sol & xhigh & max \\
\bottomrule
\end{tabular}
\end{table}

\paragraph{Episode performance.}
We measure performance by the final game score, except in \emph{Desert Racing}.
In \emph{Desert Racing}, the objective prioritizes completed laps, then checkpoint progress, with shorter completion time breaking ties.
We summarize this outcome as the number of checkpoints cleared divided by elapsed active race time, rewarding greater progress and faster completion.
We use the observed outcome of each episode directly.

\subsection{Evaluation metrics}\label{sec:experiment-measures}

We assess support for mastery through performance improvement across repeated episodes of a fixed game.
We report learning gain to measure overall improvement, and learning slope to summarize the direction and rate of performance change across all episodes.

\paragraph{Normalization.}
Revision can change the attainable performance range, so we normalize performance separately for each game and method, including the initial games.
We divide each episode outcome by the maximum value among the 90 outcomes for that game and method, pooling 3 intelligence settings, 3 runs per setting, and 10 episodes per run.

\paragraph{Learning gain.}
Following Learn2Play Bench~\citep{li2026}, we compare the early episodes~1--3 with the late episodes~8--10 of each 10-episode run.
For a given run, let $z_e$ denote the normalized performance in episode $e$.
The phase means are
\begin{equation}
  z_B=\frac{1}{3}\sum_{e=1}^{3}z_e,\qquad
  z_L=\frac{1}{3}\sum_{e=8}^{10}z_e.
\end{equation}
We compute the learning gain $\mathrm{M}_{\mathrm{gain}}$ as
\begin{equation}
  \mathrm{M}_{\mathrm{gain}}=100(z_L-z_B).
  \label{eq:learning-gain}
\end{equation}
Learning gain is measured in percentage points and captures the improvement from the early to the late phase.

\paragraph{Learning slope.}
Following Learn2Play Bench~\citep{li2026}, we fit a straight line to normalized performance as a function of episode number.
We estimate the learning slope $\mathrm{M}_{\mathrm{slope}}$ for each run by ordinary least squares with an intercept.
\begin{equation}
  \mathrm{M}_{\mathrm{slope}}=100\frac{\sum_{e=1}^{10}(e-\bar e)(z_e-\bar z)}{\sum_{e=1}^{10}(e-\bar e)^2},
  \qquad \bar e=5.5,\quad \bar z=\frac{1}{10}\sum_{e=1}^{10}z_e.
  \label{eq:learning-slope}
\end{equation}
Learning slope is measured in normalized percentage points per episode.
A positive value indicates an upward fitted trend, while a negative value indicates a downward trend.
The fit uses all 10 episode outcomes and summarizes the full-run trend.

We compute both measures for each run, then average their values across the 3 runs within each intelligence setting.
For learning slope, fitting each run and averaging the slopes gives the same result as fitting the mean performance at each episode, since all runs have the same episode indices and fitting weights.
For each intelligence setting, we compute the macro-average of the 10 per-game means.
Both measures use normalized performance, so the macro-average summarizes learning across games with different performance scales while giving each game equal weight.

\subsection{Results}

According to Table~\ref{tab:learning-gain}, PlayCraft has the largest macro-average learning gain under all 3 intelligence settings.
Under the Low setting, PlayCraft reaches a macro-average learning gain of 13.7 percentage points.
It outperforms the initial games by +2.7, with higher gains in 8 of 10 games.
It also outperforms Code by +6.3 (higher gains in 6 of 10 games) and Policy by +1.4 (higher gains in 6 of 10 games).
Under the Medium setting, PlayCraft reaches a macro-average learning gain of 31.5 percentage points.
It outperforms the initial games by +14.0, with higher gains in 9 of 10 games.
It also outperforms Code by +8.0 (higher gains in 6 of 10 games) and Policy by +15.3 (higher gains in 8 of 10 games).
Under the High setting, PlayCraft reaches a macro-average learning gain of 34.3 percentage points.
It outperforms the initial games by +21.9, with higher gains in 8 of 10 games.
It also outperforms Code by +18.2 (higher gains in 7 of 10 games) and Policy by +11.9 (higher gains in 7 of 10 games).

According to Table~\ref{tab:learning-slope}, PlayCraft has the largest macro-average learning slope under the Medium and High settings.
Under the Low setting, PlayCraft reaches a macro-average learning slope of 1.95.
It is nearly tied with the initial games (trailing by about 0.0033), and it achieves steeper slopes in 7 of 10 games.
It outperforms Code by +0.80 (steeper slopes in 6 of 10 games) and Policy by +0.13 (steeper slopes in 6 of 10 games).
Under the Medium setting, PlayCraft reaches a macro-average learning slope of 4.61.
It outperforms the initial games by +2.09, with steeper slopes in 9 of 10 games.
It also outperforms Code by +1.13 (steeper slopes in 7 of 10 games) and Policy by +1.99 (steeper slopes in 8 of 10 games).
Under the High setting, PlayCraft reaches a macro-average learning slope of 4.74.
It outperforms the initial games by +2.64, with steeper slopes in 7 of 10 games.
It also outperforms Code by +2.41 (steeper slopes in 7 of 10 games) and Policy by +1.54 (steeper slopes in 7 of 10 games).

These results reflect differences in the evidence used to guide game revision. 
Code relies on code analysis without evidence from a player agent.
Policy uses controllers that access internal game state and do not retain experience across attempts. 
PlayCraft uses an agent that pursues game objectives through the same interface as human players and learns across episodes by retaining memory and lessons. 
The agent reports the strategies it discovers, the difficulties it encounters, and how its performance changes with practice. 
This evidence helps the crafter account for existing player abilities and design challenges that reward further learning.

The results also show that learning gain and learning slope generally increase as player intelligence rises.
Players with higher intelligence may discover better strategies and execute them more effectively.
From Low to High, PlayCraft shows the largest increases in both measures.
Its learning gain rises by 20.5 percentage points, compared with 8.7 for Code, 10.0 for Policy, and 1.4 for the initial games.
Its learning slope rises by 2.78, compared with 1.17 for Code, 1.38 for Policy, and 0.14 for the initial games.
The larger increases for PlayCraft suggest that it creates more opportunities for players to improve through stronger play.

However, PlayCraft shows smaller increases from Medium to High than from Low to Medium.
For learning gain, the increase is 2.8 percentage points from Medium to High and 17.8 percentage points from Low to Medium.
For learning slope, the corresponding increases are 0.13 and 2.66.
One possible explanation is that PlayCraft uses the Medium player configuration to collect the play records that guide its revisions.
Players in the High setting can handle the resulting challenges more easily.
Over the first 3 episodes, they achieve a mean normalized performance of 47.3\%, compared with 38.2\% for players in the Medium setting. This higher starting performance may leave less room for improvement in later episodes.

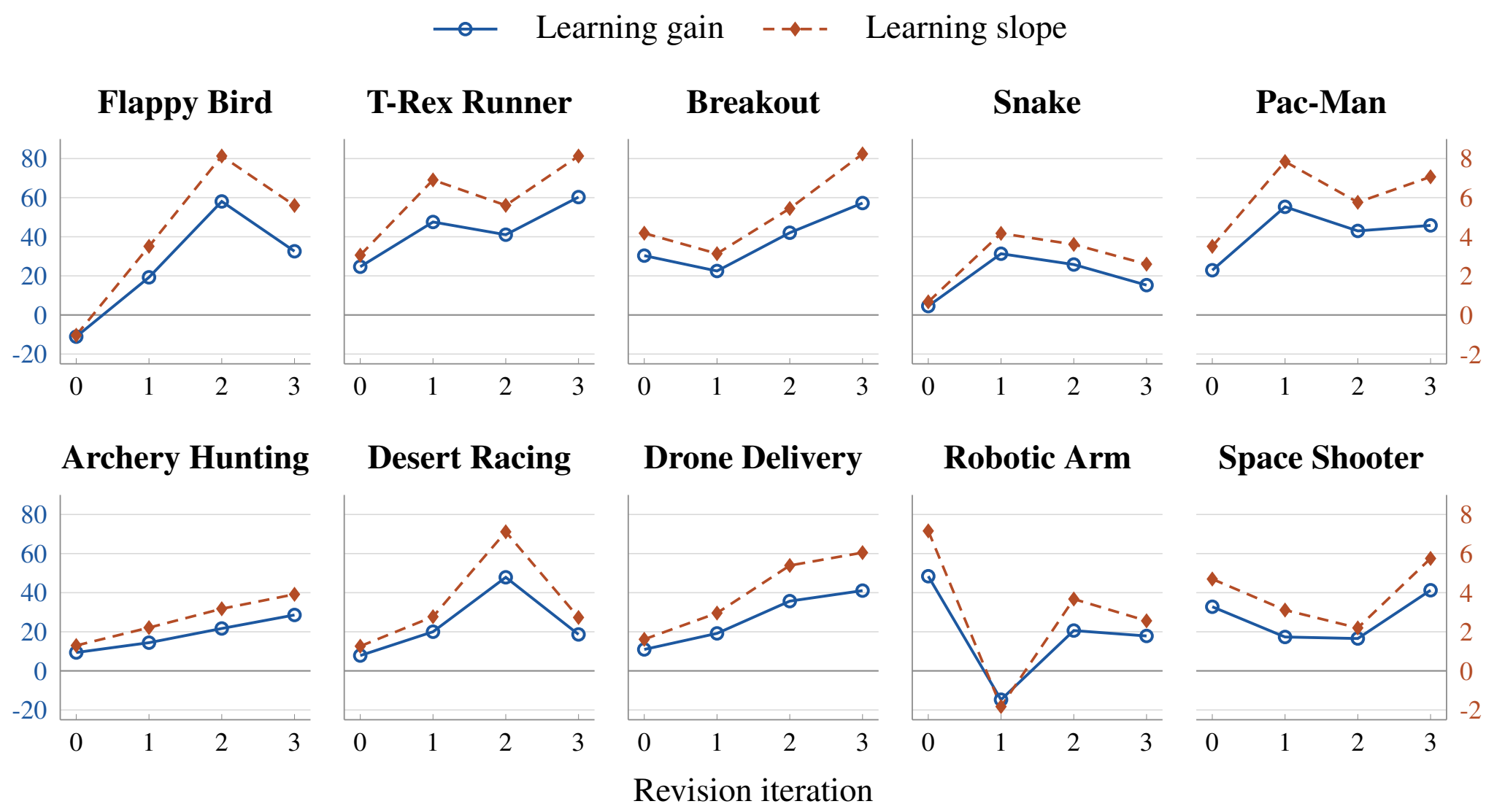
\begin{figure}
  \centering
  \resizebox{\linewidth}{!}{\begingroup
\definecolor{revisionGain}{HTML}{1D58A0}
\definecolor{revisionSlope}{HTML}{B44C26}
\pgfplotsset{
  revision gain/.style={
    revisionGain, line width=0.75pt, mark=o, mark size=1.55pt,
    mark options={solid, fill=white}
  },
  revision slope/.style={
    revisionSlope, densely dashed, line width=0.7pt, mark=diamond*, mark size=1.45pt,
    mark options={solid}
  },
  revision gain axis/.style={
    yticklabels={-20,0,20,40,60,80},
    yticklabel style={text=revisionGain}
  },
  revision slope axis/.style={
    axis y line*=right, yticklabel pos=right,
    yticklabels={-2,0,2,4,6,8},
    yticklabel style={text=revisionSlope}
  }
}
\newcommand{\revisionplots}[1]{%
  \pgfplotstableread[row sep=\\]{%
    iteration gain slope\\
    #1
  }\revisiondata
  \addplot[black!45, line width=0.4pt, forget plot] coordinates {(-0.22,0) (3.22,0)};
  \addplot[revision gain] table[x=iteration, y=gain] {\revisiondata};
  \addplot[revision slope] table[x=iteration, y expr={10*\thisrow{slope}}] {\revisiondata};
}
\begin{tikzpicture}
\begin{groupplot}[
  group style={group name=revision grid, group size=5 by 2, horizontal sep=9pt, vertical sep=35pt},
  scale only axis, width=0.168\linewidth, height=60pt,
  xmin=-0.22, xmax=3.22, ymin=-25, ymax=90,
  axis x line*=bottom, axis y line*=left,
  axis line style={black!45, line width=0.35pt},
  xtick={0,1,2,3}, xticklabels={0,1,2,3},
  ytick={-20,0,20,40,60,80}, yticklabels={},
  tick label style={font=\scriptsize},
  tick style={black!45}, major tick length=1.5pt, y tick style={draw=none},
  ymajorgrids, grid style={black!15, line width=0.3pt},
  every axis title shift=0pt,
  title style={font=\footnotesize\bfseries, anchor=base, yshift=7pt}
]
  \nextgroupplot[
    title={Flappy Bird}, revision gain axis,
    legend to name=revision legend, legend columns=2,
    legend style={draw=none, font=\footnotesize, column sep=8pt}
  ]
  \revisionplots{
    0 -11.111111 -1.037037\\
    1  19.259259  3.510662\\
    2  58.080808  8.131313\\
    3  32.575758  5.601469\\
  }
  \legend{{Learning gain}, {Learning slope}}

  \nextgroupplot[title={T-Rex Runner}]
  \revisionplots{
    0  24.668561  3.059286\\
    1  47.567568  6.907999\\
    2  41.051051  5.607426\\
    3  60.300300  8.129948\\
  }

  \nextgroupplot[title={Breakout}]
  \revisionplots{
    0  30.352304  4.183296\\
    1  22.486772  3.136123\\
    2  42.063492  5.447330\\
    3  57.246377  8.238911\\
  }

  \nextgroupplot[title={Snake}]
  \revisionplots{
    0   4.545455  0.670340\\
    1  31.327160  4.173681\\
    2  25.800377  3.608971\\
    3  15.259259  2.599327\\
  }

  \nextgroupplot[title={Pac-Man}, revision slope axis]
  \revisionplots{
    0  22.881944  3.507576\\
    1  55.316013  7.846960\\
    2  42.994043  5.762991\\
    3  45.816993  7.067142\\
  }

  \nextgroupplot[title={Archery Hunting}, revision gain axis]
  \revisionplots{
    0   9.391006  1.296794\\
    1  14.421584  2.211703\\
    2  21.680083  3.177114\\
    3  28.593188  3.916547\\
  }

  \nextgroupplot[title={Desert Racing}]
  \revisionplots{
    0   7.837820  1.261865\\
    1  20.014491  2.770505\\
    2  47.871342  7.115055\\
    3  18.632900  2.730674\\
  }

  \nextgroupplot[title={Drone Delivery}]
  \revisionplots{
    0  10.962666  1.614947\\
    1  19.175241  2.957790\\
    2  35.680263  5.390774\\
    3  41.058833  6.048989\\
  }

  \nextgroupplot[title={Robotic Arm}]
  \revisionplots{
    0  48.419980  7.159670\\
    1 -14.747475 -1.826840\\
    2  20.619439  3.674891\\
    3  17.855340  2.560591\\
  }

  \nextgroupplot[title={Space Shooter}, revision slope axis]
  \revisionplots{
    0  32.799411  4.698183\\
    1  17.334096  3.111319\\
    2  16.547268  2.197576\\
    3  41.245504  5.751553\\
  }
\end{groupplot}
\node[anchor=south, yshift=18pt] at (revision grid c3r1.north) {\pgfplotslegendfromname{revision legend}};
\node[anchor=north, yshift=-12pt, font=\footnotesize] at (revision grid c3r2.south) {Revision iteration};
\end{tikzpicture}
\endgroup}
  \caption{Learning gain and learning slope across PlayCraft revision iterations. Learning gain uses the left y-axis scale, and learning slope uses the right y-axis scale. Iteration 0 is the initial game. For each game and iteration, performance is normalized by the largest observed outcome. Each point averages three Medium runs.}
  \label{fig:revision-iterations}
\end{figure}

Figure~\ref{fig:revision-iterations} tracks learning gain and learning slope from each initial game through 3 successive PlayCraft revisions.
Both measures generally increase across revisions, with fluctuations that may reflect variability in gameplay, diagnosis, design, or implementation.
After the final revision, both measures exceed their initial values in 9 of the 10 games.
Both measures drop most sharply in \emph{Robotic Arm} after the first revision and partially recover after the second.
The first revision introduced wrist sway under load and varied the starting positions of the cans, which may have made transfers harder to execute consistently. Section~\ref{sec:qualitative-analysis} discusses these changes to \emph{Robotic Arm} in more detail.

\begin{table}[!t]
\centering
\footnotesize
\setlength{\tabcolsep}{2pt}
\renewcommand{\arraystretch}{1.08}
\caption{Learning gain. Values are averaged over 3 runs at each intelligence setting. Bold marks the largest mean in each row within each setting.}
\label{tab:learning-gain}
\resizebox{\linewidth}{!}{%
\begin{tabular}{@{}lrrrr@{\hspace{4pt}}|@{\hspace{4pt}}rrrr@{\hspace{4pt}}|@{\hspace{4pt}}rrrr@{}}
\toprule
 & \multicolumn{4}{c|@{\hspace{4pt}}}{Low} & \multicolumn{4}{c|@{\hspace{4pt}}}{Medium} & \multicolumn{4}{c}{High} \\
\cmidrule(lr){2-5}\cmidrule(lr){6-9}\cmidrule(lr){10-13}
Game & Initial & Code & Policy & PlayCraft & Initial & Code & Policy & PlayCraft & Initial & Code & Policy & PlayCraft \\
\midrule
\emph{Flappy Bird} & 1.4 & 0.7 & 3.8 & \textbf{15.7} & -3.5 & \textbf{38.8} & 19.4 & 32.6 & 16.3 & 4.7 & \textbf{40.4} & 23.7 \\[3pt]
\emph{T-Rex Runner} & \textbf{38.6} & 23.3 & 36.1 & 15.3 & 24.7 & 45.2 & 25.4 & \textbf{60.3} & -6.0 & 37.3 & 16.4 & \textbf{70.3} \\[3pt]
\emph{Breakout} & -9.6 & -19.1 & 4.2 & \textbf{8.4} & 24.9 & 11.8 & 19.8 & \textbf{52.7} & 13.8 & 10.7 & 29.3 & \textbf{48.2} \\[3pt]
\emph{Snake} & -6.3 & 6.6 & 3.8 & \textbf{8.0} & 4.3 & \textbf{15.0} & -0.6 & 13.0 & -1.9 & 8.7 & 7.3 & \textbf{26.7} \\[3pt]
\emph{Pac-Man} & 18.6 & 13.8 & 24.5 & \textbf{31.0} & 20.9 & \textbf{35.8} & 25.2 & 22.2 & 22.6 & 9.7 & \textbf{43.9} & 3.1 \\[3pt]
\emph{Archery Hunting} & 5.7 & -0.9 & 3.8 & \textbf{11.6} & 8.6 & 9.4 & 14.3 & \textbf{27.0} & 10.6 & 15.8 & 8.3 & \textbf{29.8} \\[3pt]
\emph{Desert Racing} & 17.7 & \textbf{25.8} & 10.3 & 17.9 & 7.8 & 11.1 & 9.4 & \textbf{18.6} & 1.6 & 13.5 & \textbf{25.7} & 11.9 \\[3pt]
\emph{Robotic Arm} & 17.7 & 2.9 & \textbf{20.7} & 20.4 & 48.4 & \textbf{54.2} & 24.0 & 16.1 & \textbf{47.2} & 22.2 & 17.3 & 20.2 \\[3pt]
\emph{Drone Delivery} & \textbf{28.8} & 17.0 & 12.5 & 9.9 & 10.9 & 4.8 & 27.7 & \textbf{40.0} & 7.5 & 4.8 & 21.1 & \textbf{48.6} \\[3pt]
\emph{Space Shooter} & -2.4 & \textbf{3.9} & 3.8 & -0.6 & 27.9 & 9.0 & -2.2 & \textbf{32.7} & 12.6 & 33.7 & 13.8 & \textbf{60.3} \\[3pt]
\midrule
Macro-average & 11.0 & 7.4 & 12.3 & \textbf{13.7} & 17.5 & 23.5 & 16.2 & \textbf{31.5} & 12.4 & 16.1 & 22.3 & \textbf{34.3} \\
\bottomrule
\end{tabular}%
}
\end{table}

\begin{table}[!t]
\centering
\footnotesize
\setlength{\tabcolsep}{2pt}
\renewcommand{\arraystretch}{1.08}
\caption{Learning slope. Values are averaged over 3 runs at each intelligence setting. Bold marks the largest mean in each row within each setting.}
\label{tab:learning-slope}
\resizebox{\linewidth}{!}{%
\begin{tabular}{@{}lrrrr@{\hspace{4pt}}|@{\hspace{4pt}}rrrr@{\hspace{4pt}}|@{\hspace{4pt}}rrrr@{}}
\toprule
 & \multicolumn{4}{c|@{\hspace{4pt}}}{Low} & \multicolumn{4}{c|@{\hspace{4pt}}}{Medium} & \multicolumn{4}{c}{High} \\
\cmidrule(lr){2-5}\cmidrule(lr){6-9}\cmidrule(lr){10-13}
Game & Initial & Code & Policy & PlayCraft & Initial & Code & Policy & PlayCraft & Initial & Code & Policy & PlayCraft \\
\midrule
\emph{Flappy Bird} & 0.37 & 0.22 & 0.54 & \textbf{2.25} & -0.33 & 5.47 & 3.59 & \textbf{5.60} & 3.33 & 0.15 & \textbf{5.27} & 3.06 \\[3pt]
\emph{T-Rex Runner} & \textbf{6.11} & 3.04 & 5.73 & 2.16 & 3.06 & 5.87 & 4.48 & \textbf{8.13} & -0.50 & 5.18 & 1.76 & \textbf{9.83} \\[3pt]
\emph{Breakout} & -0.65 & -2.03 & 0.56 & \textbf{1.08} & 3.43 & 1.81 & 2.75 & \textbf{7.58} & 1.99 & 2.02 & 4.90 & \textbf{7.03} \\[3pt]
\emph{Snake} & 0.59 & 0.78 & 0.56 & \textbf{1.02} & 0.64 & \textbf{2.82} & -0.07 & 2.22 & -0.32 & 1.75 & 0.86 & \textbf{3.59} \\[3pt]
\emph{Pac-Man} & 2.76 & 1.87 & 3.11 & \textbf{4.23} & 3.21 & \textbf{5.18} & 4.04 & 3.42 & 3.20 & 1.47 & \textbf{6.19} & 0.44 \\[3pt]
\emph{Archery Hunting} & 0.57 & -0.14 & 0.87 & \textbf{1.51} & 1.19 & 1.31 & 2.03 & \textbf{3.69} & 1.53 & 2.23 & 1.31 & \textbf{4.22} \\[3pt]
\emph{Desert Racing} & 2.58 & \textbf{3.80} & 1.45 & 2.75 & 1.25 & 1.53 & 1.31 & \textbf{2.73} & 0.27 & 2.07 & \textbf{3.56} & 1.53 \\[3pt]
\emph{Robotic Arm} & \textbf{3.54} & 0.85 & 3.06 & 3.02 & 7.16 & \textbf{8.45} & 3.72 & 2.30 & \textbf{7.23} & 3.06 & 2.94 & 2.99 \\[3pt]
\emph{Drone Delivery} & \textbf{4.07} & 2.68 & 2.08 & 1.52 & 1.60 & 0.73 & 4.36 & \textbf{5.89} & 1.16 & 0.73 & 3.38 & \textbf{6.29} \\[3pt]
\emph{Space Shooter} & -0.37 & \textbf{0.47} & 0.23 & 0.01 & 3.99 & 1.68 & -0.03 & \textbf{4.55} & 3.13 & 4.59 & 1.82 & \textbf{8.41} \\[3pt]
\midrule
Macro-average & \textbf{1.96} & 1.15 & 1.82 & 1.95 & 2.52 & 3.49 & 2.62 & \textbf{4.61} & 2.10 & 2.32 & 3.20 & \textbf{4.74} \\
\bottomrule
\end{tabular}%
}
\end{table}

\subsection{Qualitative analysis}\label{sec:qualitative-analysis}

We select 4 games with different demands on planning and execution to understand how each method responds to the skills players must develop.
The cases span rebound control in \emph{Breakout}, ballistic aiming and target approach in \emph{Archery Hunting}, object manipulation in \emph{Robotic Arm}, and route planning and flight under a battery budget in \emph{Drone Delivery}.
We use the play and revision records to explain how the evidence available to each method shaped its design choices.

\paragraph{\emph{Breakout}.}

The initial game requires players to move the paddle to bounce the ball toward the bricks. A brick breaks after a single hit.
The PlayCraft player agent already aimed rebounds deliberately, but clearing isolated bricks required precise follow-up shots.
In the first round, PlayCraft introduced armor that retains damage from underside hits, so one shot can prepare a brick for a later shot.
In subsequent reports, the player described shots that struck a different column than intended and sequences of hits that cleared few bricks.
In the second round, PlayCraft slowed paddle movement for short taps and increased the number of underside hits needed to break a brick to three.
A hit on the side or top could now also clear a contiguous run of damaged neighbors in the same row.
In the third round, PlayCraft made the rebound angle less sensitive to small changes in where the ball strikes near the paddle center.
The ball could now pass through a brick on the third underside hit.
The player can prepare either a direct path through the wall or a sweep along it.

Code analyzed the collision rules and revised the paddle mechanics so that stopping the paddle before a catch widens its cutting area. 
A catch within that area can send the ball through bricks at the base of the wall and into a central passage.
Policy used controller trajectories to identify useful paths into the wall from above and lengthy return trips.
In that revision, breaking a brick from above weakens the remaining bricks in the column below.
The ball continues along its path whenever it destroys a brick.

In Code, players learned to stop the paddle before catches to widen the cutting area, producing substantial progress early.
Later practice focused on steering and positioning for returning balls, but missed catches sometimes prevented these refinements from producing higher scores.
In Policy, players learned to reach exposed tops and sides and exploit weakened columns, but prediction errors after collisions continued to cause missed catches.
At Medium and High, PlayCraft players progressed from safe, near-vertical returns that opened columns to directing shots through those openings for larger clears.
At Low, improvement mainly came from earlier interceptions and controlled column clearing.
Better interception and controlled clearing at Low, together with more effective use of prepared openings at Medium and High, may explain why PlayCraft led on learning gain and slope at all three settings.

\paragraph{\emph{Archery Hunting}.}

The initial game requires players to approach targets, choose firing positions, and adjust their aim and draw to account for gravity and wind.
The PlayCraft player agent completed the hunt by closing in on targets and shooting at full draw.
In the first round, PlayCraft added cover and made targets retreat in response to noise from movement, bow releases, and ordinary hits.
A quiet hit on a gold weak spot could leave nearby targets undisturbed.
In the second round, PlayCraft let the player flush targets from one refuge toward the other and extended their exposure during crossings to four seconds.
The player returned to full-draw shots after a quiet shot missed and a prolonged pursuit stalled.
The crafter found that pursuit could still work because targets crossed exposed ground between refuges and ignored new disturbances while retreating.
In the third round, PlayCraft moved the approaches to the escape openings under cover.
When a retreating target reaches a fork, it chooses an escape route based on the latest disturbance it detected.
A quiet hit on a gold weak spot can leave the next target on course. Before a target reaches its fork, a deliberate disturbance can steer it toward an opening where the player can intercept it.
Feedback on ordinary hits now shows where the arrow lands relative to the gold weak spot.

Code used ballistic probes to find that, at close range, full-draw shots could hit even alerted targets without leading them.
It added fallen trunks so players have a reason to arc a lower-draw shot over cover or move to the side for a direct shot.
Policy controllers often moved between firing positions or withheld shots while looking for a clear firing window.
The revision added shared firing areas from which the player can target several marks.
Stone screens limit which targets hear an impact, giving the player more control over which targets retreat.

In Code, players completed the hunt from the first episode, mostly by moving around cover and using strong draws, then refined their aim for weak-spot hits.
In Policy, players often used quiet draws and waited or repositioned for clearer target faces, reducing speculative shots and conserving arrows.
Both baselines often reached completion early, so later practice mainly improved shot precision and efficiency.
In PlayCraft, brief shooting opportunities and targets retreating behind cover led to missed shots and prolonged searches, often leaving early hunts incomplete.
Players learned to abandon unproductive pursuits, prepare shots for clearer openings, and use directional hit feedback to refine weak-spot aim.
Progress toward completing the hunt, together with improved shot precision, may explain why PlayCraft led on learning gain and slope at all three settings.

\paragraph{\emph{Robotic Arm}.}

The initial game requires players to align the gripper, carry each can to the receiver, and release it so it settles on the surface.
The PlayCraft player agent could already transfer cans quickly by using separate landing areas.
In the first round, PlayCraft added wrist sway under load and varied the starting positions of the cans.
In the second round, PlayCraft introduced a rotating receiver.
Delivered cans move with the deck and remain physical obstacles.
As a result, the timing and position of one delivery can affect clearance on the next.
In the third round, PlayCraft revised contact with the receiver after the player reported difficulty withdrawing the gripper and judging whether the deck supported the can.
The deck rotates more slowly when the player rests a held can gently on it.
The player can brace the deck until it stops or release the can while the deck is still moving.

Code used clearance analysis and timed probes to preserve two ways of reaching a stationary receiver.
The player can carry a can through a low opening or over the rail.
The rail, the different can heights, and the ability to tilt the wrist make grip and posture important when clearing obstacles.
Policy controllers struggled to pick up cans, so the final revision moved the obstructing bridges behind the gripping pads, leaving the approach to the claw open.
The revision also lets the player guide a can into place while the stationary receiver supports its weight.

In Code, players learned to adjust grip depth and wrist posture to carry cans over the fixed rail and repeat successful interior placements.
In Policy, players learned to lower cans well inside the stationary receiver and coordinate release, settling, and claw withdrawal.
In both baselines, these strategies generally produced repeatable completions, allowing further gains from shorter transfers and pauses.
PlayCraft players also learned to place cans farther inside the deck and leave clearance for withdrawal, but missed grips, rim falls, and interference from rotating cans continued after earlier fast completions.
Persistent correction and recovery delays reduced completion bonuses even after players had learned successful placements.
This inconsistency may explain why PlayCraft trailed at least one baseline on both learning metrics at every intelligence setting.

PlayCraft also yielded lower learning gain and slope than the initial game, as shown in Figure~\ref{fig:revision-iterations}.
During the play run preceding the first revision, the agent quickly learned a conservative transfer routine and repeated it reliably.
The first revision added wrist sway and varied the starting positions of the cans, requiring adjustments to braking and pickup approaches.
These additional demands may have made learned routines harder to repeat reliably, which may explain the sharp decline in learning gain and slope after the first revision.
Later revisions partially recovered both measures, but they remained below their initial values in that comparison.

\paragraph{\emph{Drone Delivery}.}

The initial game requires players to collect and deliver packages in order, choose flight paths that conserve battery power, and slow down as they approach each pad.
The PlayCraft player agent quickly settled on a clean route.
In the first round, PlayCraft introduced wind at higher altitudes. The player found a repeatable route through the calm air along the streets.
In the second round, PlayCraft limited shelter to areas behind individual buildings.
The player reported that prolonged coasting could stall progress. The crafter traced this to a speed limit in the code that could abruptly slow the drone on entering a headwind.
The player also completed two deliveries with less damage than in the preceding attempt, but earned the same score.
In the third round, PlayCraft let the drone carry its momentum into a headwind, with drag slowing it gradually.
The revision also awards bonuses for remaining battery power and low damage at each delivery.
Players can build speed in shelter and coast across an exposed stretch.
They can also earn credit for keeping the drone in better condition even if they complete only part of the route.

Code used source analysis and route probes to develop a different trade-off.
Carrying two packages can reduce the number of trips, and delivering while moving can reduce the number of stops.
The extra load makes the drone slower to respond.
Policy controllers spent time waiting for favorable wind or attempting to travel slowly along the ground.
They reached the time cutoff with substantial battery power remaining.
The final revision added runners to the suspended load so the player can tow it more quickly between flights by keeping it aligned on packed snow.

In Code, players adopted batching early and usually used hovering deliveries, learning to coast with heavier loads through open corridors and brake early near obstacles and pads. Once a safe route became reliable, further practice mainly refined pad alignment and produced only small gains.
In Policy, players learned to align the runners, settle the load before turning, and balance cargo preservation against battery use, often abandoning airborne transport after severe cargo damage.
Protecting the more valuable later loads remained difficult, but better handling increased scores even after players could complete the route.
PlayCraft encouraged players to build speed in shelter, coast across exposed gaps, and brake accurately while conserving enough battery for all three deliveries.
At Medium and High, practice produced more reliable routes that reached later deliveries and earned their larger bonuses, which may explain why PlayCraft led on learning gain and slope.
At Low, players described similar energy-saving strategies but usually failed to reach the second delivery because of battery exhaustion or collisions.
This persistent difficulty in execution may explain why PlayCraft trailed the initial game and both baselines on both learning metrics.

\section{Conclusion}\label{sec:conclusion}

We presented PlayCraft, an automated framework for making games more engaging by supporting mastery.
A player agent interacts through the same interface as human players, observing the game screen and using the controls to pursue game objectives. 
It plays multiple episodes, reviewing its play after each one and carrying its memory from one episode to the next.
PlayCraft uses this play experience to identify gaps between expected and target performance at different proficiency levels and guide changes that enable player performance gain through better decisions and execution.
At each player intelligence setting, fresh evaluation runs showed larger learning gains and steeper learning slopes in most games revised by PlayCraft than in their initial counterparts.
PlayCraft also outperformed the baselines on the macro-average of each metric at every setting.

\textbf{Limitations.} PlayCraft assumes that the play and learning patterns of agents resemble those of human players.
Training on human-generated data may help foundation models approximate human decisions, but this assumption remains unverified.
Future work should align agent play with human behavior and test that alignment by comparing the decisions, difficulties, and learning patterns of agents and human players.
Confirming the alignment would make agent play evidence more reliable for game crafting and evaluation.
The second limitation is the cost of playing games that require longer runs.
Increasing the active play budget can require more gameplay time and model inference.
More efficient observation and action planning could make longer runs practical and extend evaluation to a broader range of games.

\FloatBarrier
\bibliographystyle{conference}
\bibliography{references}

\end{document}